\documentclass[10pt,journal,compsoc]{IEEEtran}
\IEEEoverridecommandlockouts
\ifCLASSOPTIONcompsoc
  \usepackage[nocompress]{cite}
\else
  \usepackage{cite}
\fi
\usepackage{authblk}
\usepackage{amsmath,amssymb,amsfonts}
\usepackage{algorithmic}
\usepackage[titlenumbered, ruled, linesnumbered, vlined]{algorithm2e}
\usepackage{graphicx}
\usepackage{caption,subcaption}
\usepackage{textcomp}
\usepackage{xcolor, colortbl}
\usepackage{todonotes,soul}
\usepackage{multirow}
\usepackage{url}
\usepackage{hyperref}
\usepackage{booktabs}
\usepackage{calc}
\def\BibTeX{{\rm B\kern-.05em{\sc i\kern-.025em b}\kern-.08em
    T\kern-.1667em\lower.7ex\hbox{E}\kern-.125emX}}
\usepackage{soul}
\soulregister\cite7
\soulregister\ref7
\soulregister\pageref7

\def\fixme#1{\typeout{FIXED in page \thepage : {#1}}
\bgroup \color{red}{[FIXME: {#1}]} \egroup}

\hypersetup{
    colorlinks=true,
    linkcolor=blue,
    urlcolor=blue, 
    citecolor=blue, 
}

\begin{document}

\title{On Exploring Input Resolution Scaling For Anytime LiDAR Object Detection}

\author{Ahmet~Soyyigit,
        Shuochao~Yao,
        and~Heechul~Yun%
\thanks{Dr. Soyyigit is with The National Defense University, Istanbul, Turkiye (e-mail: ahmet.soyyigit@msu.edu.tr). He is the corresponding author of this paper.}%
\thanks{Dr. Yao is with George Mason University, Fairfax, VA, USA (e-mail: shuochao@gmu.edu).}%
\thanks{Dr. Yun is with The University of Kansas, Lawrence, KS, USA (e-mail: heechul.yun@ku.edu).}%
% \thanks{Manuscript received February 5, 2026; revised Month DD, YYYY.}
}

\maketitle

\begin{abstract}
Making tradeoffs between execution latency and result utility 
(i.e., anytime computing) for adapting to dynamic operational 
requirements has been shown to enhance the performance of 
cyber-physical systems. In this work, we focus on enabling anytime 
computing for deep neural networks (DNNs) that process LiDAR point 
clouds for 3D object detection. We propose a novel method that enables
multi-resolution inference for models that process point clouds as
pillars or voxels, allowing the input to be dynamically scaled and 
processed at the resolution needed to meet timing requirements. 
Importantly, our memory-efficient approach requires the deployment of 
only a single DNN model, avoiding the need to deploy multiple models,
each trained for a different input resolution. We also introduce a 
deadline-aware scheduler that selects the highest possible resolution
for any given input by accurately predicting the execution time for
all possible resolutions at runtime, which is challenging due to 
the irregularity of LiDAR point clouds. Experimental results on 
the nuScenes autonomous driving dataset demonstrate that our method
significantly outperforms existing anytime computing approaches 
for LiDAR object detection. Finally, we deploy our approach in a
simulated autonomous driving system, where it consistently enables
collision-free navigation while avoiding unnecessary stalls caused
by environmental complexity.
\end{abstract}

\begin{IEEEkeywords}
LiDAR, 3D object detection, Deep neural networks, Anytime computing, Simulation
\vspace{-10pt}
\end{IEEEkeywords}

\section{Introduction}
\label{sec:introduction}

Autonomous systems are critically dependent on the accurate 
detection of surrounding objects in real-time.
For this task, numerous highly accurate LiDAR-based object detection 
deep neural networks (DNNs) have been proposed in recent 
years~\cite{lidaro, centerpoint, pillarnet, pointpillars}.
However, these state-of-the-art LiDAR object detection DNNs are 
computationally expensive, making deployment on resource-constrained 
embedded computing hardware challenging.
This challenge is particularly pronounced in systems with strict size, 
weight, and power (SWaP) constraints, necessitating trade-offs between 
accuracy and latency.

The required accuracy/latency trade-offs depend not only on the SWaP 
constraints but also on the dynamic operation 
environment~\cite{gog2022ctxaware, d3}.
For example, in complex and crowded urban environments where objects 
move slowly, processing input in a fine-grained manner may be 
desirable to maximize detection accuracy, even if it takes longer.
However, in simpler environments with fast-moving objects, 
such as highways, it may be preferable to process quickly in a 
coarse-grained manner, as lower processing latency could be more 
important than high precision and fine-grained details.

Algorithms that can trade off quality and latency are known as \textit{anytime} algorithms in the literature, and there has been significant effort in recent years to make anytime-capable DNNs that process perceptual input data.
For image classification and object detection tasks, ``early-exit'' architectures have been explored~\cite{kuhse2025anytimeyolo, yao2020imprecisecomp, liu2022rttasksched, kim2020abc}, where additional output layers are integrated at intermediate stages of a DNN to allow predictions to be made before reaching the full depth of the model.
Criticality-based slicing and scheduling of input~\cite{hu2021exploring,liu2022rttasksched,liu2020removing,kang2022dnnsam,rtbev} and dynamic scaling of image resolution~\cite{heo2022rtscale,rsnets,zhu2021dynamicresolutionnetwork} have been studied to enable anytime processing capabilities in object classification and detection DNNs.
However, most prior works have focused on DNNs that process camera images.

For LiDAR-based object detection tasks, Anytime-LiDAR~\cite{alidar} combined the early-exit method with a novel detection head scheduling technique to enable dynamic latency/accuracy trade-offs for PointPillars~\cite{pointpillars}. 
VALO~\cite{VALO2024} explores a deadline-aware input slicing and scheduling approach that greatly improves anytime performance, achieving higher accuracy across a gamut of deadlines, when applied to the state-of-the-art LiDAR object detection models~\cite{centerpoint,voxelnext}.
However, input resolution, defined here as the spatial granularity of the input encoded from the LiDAR scans, remains a largely underexplored scaling factor in the design of anytime LiDAR detection models.
Although adjusting resolution provides an excellent trade-off between detection accuracy and execution time (as shown in Figure~\ref{fig:e2eboxplot}), the runtime memory requirements grow linearly with the number of supported resolutions due to the need for a separate model for each resolution, presenting a key challenge for practical deployment.

\begin{figure*}[t]
\centerline{\includegraphics[scale=0.65]{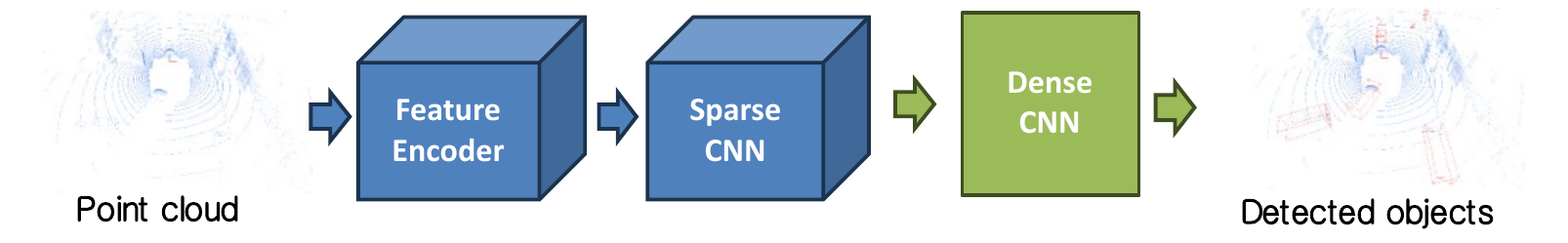}}
\caption{General architecture of LiDAR Object Detection DNNs.}
\label{fig:pillarnet}
\end{figure*}

%.%.%.

In this paper, we propose MURAL~\footnote{This work extends a prior conference publication~\cite{mural2025} by adding voxel-based DNN support and closed-loop evaluation.}, a multi-resolution anytime framework 
for LiDAR object detection DNNs.
First, MURAL enables dynamic selection of input resolution, 
allowing flexible trade-offs between accuracy and latency while using 
a single shared set of network weights.
This is possible thanks to its multi-resolution architecture 
enhancement and training methodology (Section~\ref{sec:multires_training}).
Second, MURAL can support additional input resolutions on top of those used during training, by synthesizing 
new input resolutions
(Section~\ref{sec:posttrainingpillars}).
Third, MURAL incorporates a deadline-aware scheduler that dynamically
selects the highest feasible input resolution for a given
time constraint, based on accurate execution time predictions
for each resolution (Section~\ref{sec:dl_aware_sched}).

We evaluate MURAL on two 2D pillar-based models: the state-of-the-art 
Pillarnet \cite{pillarnet} and the widely used PointPillars 
\cite{pointpillars}. To demonstrate the versatility of our approach, 
we further apply MURAL to CenterPoint \cite{centerpoint}, a popular 3D 
voxel-based DNN. We assess performance in terms of accuracy, latency, 
and resource utilization, comparing MURAL against separately trained 
baseline models for each resolution as well as the prior 
state-of-the-art anytime method~\cite{VALO2024}.

Our evaluation is two-fold: (i) a hard-deadline, open-loop setting 
using a large-scale autonomous driving dataset, and (ii) a closed-loop 
simulated environment in which object detection directly influences 
navigation performance. Results show that MURAL achieves 
higher detection accuracy under a wide range of deadlines compared 
to both the baselines and the prior anytime approach. In the 
closed-loop experiments, we demonstrate that MURAL improves 
navigation safety and efficiency by adjusting input resolution 
according to the ego-vehicle’s velocity, outperforming 
fixed-resolution baselines.

In summary, our key contributions are:
\begin{itemize}
\item We present the first work enabling runtime resolution 
scaling for LiDAR-based object detection DNNs and release it open-source~\footnote{MURAL code repository: \url{https://github.com/CSL-KU/MURAL}}.
\item We introduce a general framework applicable to 
both pillar- and voxel-based detection architectures.
\item We achieve superior accuracy–latency trade-offs compared to 
non-anytime baseline models and the prior state-of-the-art anytime 
LiDAR approach.
\item We demonstrate the practical benefits of MURAL in a
closed-loop autonomous driving simulation, showing improved navigation 
safety and efficiency.
\end{itemize}

\section{Background}
\label{sec:background}

\begin{figure}[htp]
\centerline{\includegraphics[scale=0.45]{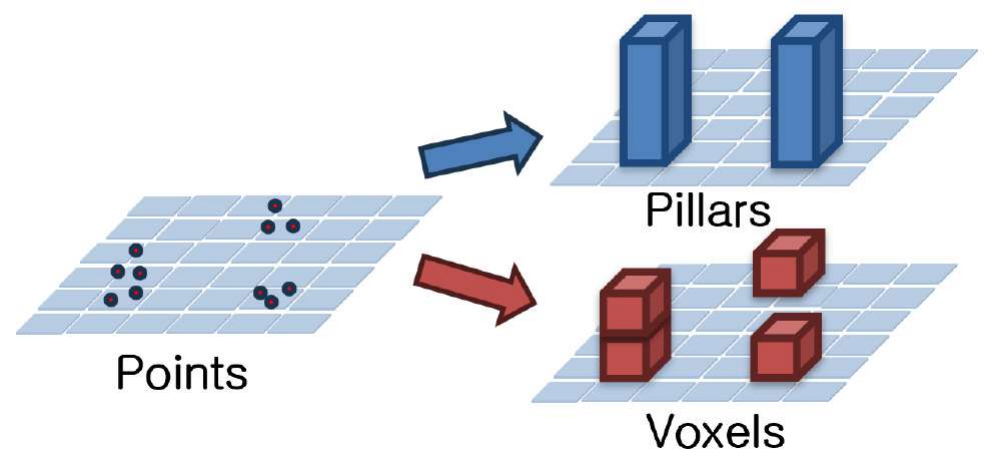}}
\caption{3D points converted to voxels and pillars.}
\label{fig:pts_pillar_vox}
\end{figure}

\begin{figure*}[t]
\centerline{\includegraphics[scale=0.42]{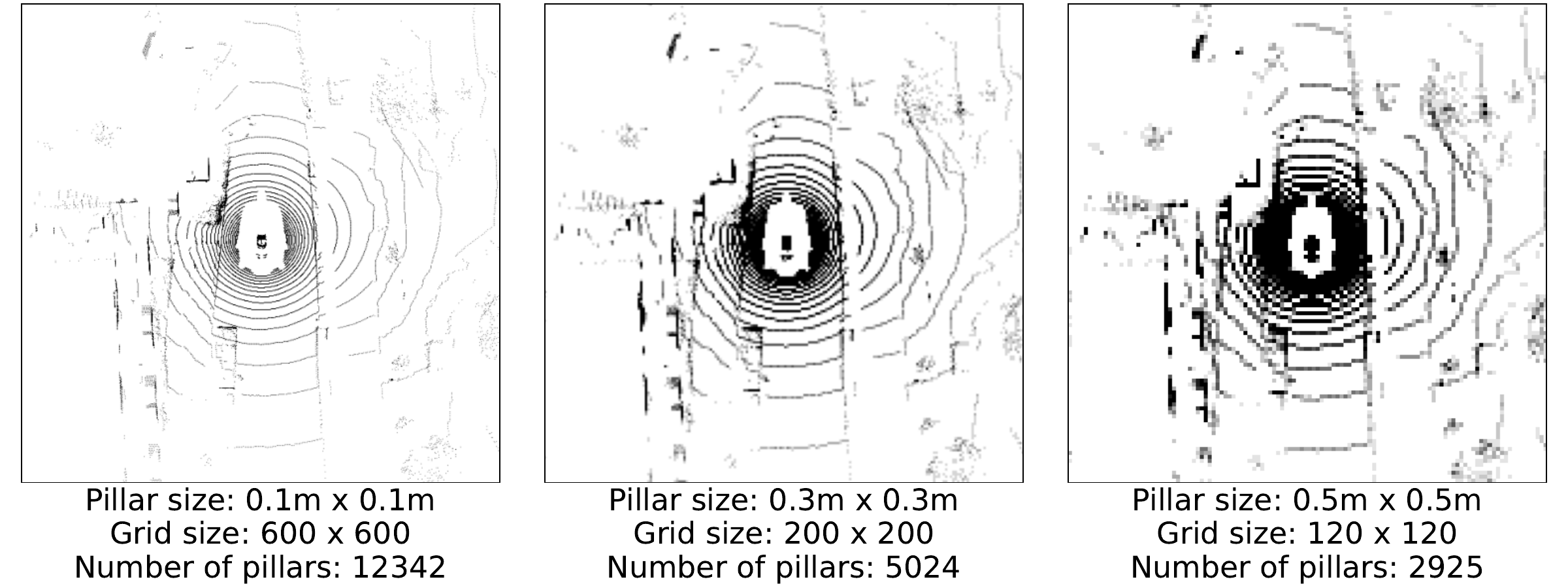}}

\caption{Bird's-eye-views of a LiDAR point cloud transformed into pillars of three sizes. Darkness indicates point density.
%In all cases, the cubical space $S$ that contains the point cloud (thus pillars) is defined by range $(X_s=-30m, X_e=30m, Y_s=-30m, Y_e=30m, Z_s=0m, Z_e=8m)$.
%A darker color indicates having more points in a pillar.
}
% \vspace{-10pt}
\label{fig:pillars}
\end{figure*}

In this section, we provide the necessary background on LiDAR object detection DNNs and resolution scaling.

\subsection{LiDAR Object Detection DNNs}

The LiDAR sensor continuously scans the surrounding environment, producing periodic snapshots. Each snapshot is represented as a point cloud \textit{P} consisting of \textit{n} points, formally defined as:
\begin{equation}
P = \{(x_1, y_1, z_1, i_1), \ldots, (x_n, y_n, z_n, i_n)\}
\end{equation}
where each point encodes its 3D spatial coordinates in meters along with the laser return intensity.
DNNs have emerged as a powerful approach for detecting objects of interest within point clouds \cite{second,pointpillars,centerpoint,pillarnet}.
To enable efficient DNN-based processing, the cubic space $S$ encompassing the point cloud is partitioned into a grid $G$ of uniformly sized cubical cells.
Cells that contain at least one point are referred to as \textit{voxels}.
The dimensions of $G$ are determined by:
\begin{equation}
\label{grideq}
    G = (\frac{X_e - X_s}{V_x}, \frac{Y_e - Y_s}{V_y}, \frac{Z_e - Z_s}{V_z})
\end{equation}
where $(X_s, X_e, Y_s, Y_e, Z_s, Z_e)$ define the range of $S$ in the LiDAR-centered coordinate system and $(V_x, V_y, V_z)$ denotes the voxel size, both expressed in meters.
Converting a point cloud into a voxel representation enables the application of convolutional neural networks (CNNs) for feature extraction, as the resulting voxel grid can be treated as either a sparse or dense tensor (i.e., a multidimensional array).

% The points are mapped to their voxel coordinates in $G$ as follows:
% \begin{equation}
%     x' = \frac{(x - X_s)}{V_x}, \quad y' = \frac{(y - Y_s)}{V_y}, \quad z' = \frac{(z - Z_s)}{V_z}
% \end{equation}

When the height of the voxels $(V_z)$ is equal to the height of the cubic space $(Z_e - Z_s)$, effectively removing the height dimension of $G$, the voxels are called instead \textit{pillars}.
Figure~\ref{fig:pts_pillar_vox} illustrates the conversion of raw points into both pillars and voxels.
In practice, a DNN typically employs only 
one of these representations, as they require distinct feature encoders
and specialized model architectures. Several works have proposed using 
pillars rather than voxels~\cite{pointpillars,pillarnet} to avoid 
computationally expensive 3D convolutional layers, thereby providing 
deployment-friendly solutions with minimal sacrifice in detection 
accuracy.

Figure~\ref{fig:pillarnet} illustrates the general architecture of
LiDAR-based object detection DNNs. The process begins with a feature
encoder (FE) that transforms raw points into pillars or voxels,
typically represented in a coordinate list (COO) format. At this stage,
the sparse data occupy only a small fraction of the grid
(e.g., 3\%--20\%). Consequently, converting this sparse tensor into a dense tensor and applying standard dense convolutions—as used in image
processing—would be computationally wasteful.

To address this, sparse CNNs~\cite{second} are employed to process the
data directly in its sparse format. Sparse convolutions apply the same mathematical operations as dense convolutions, but restrict computation to non-zero (occupied) elements, skipping empty locations to reduce computational cost.
Output of the sparse CNN is scattered onto a dense grid of zeros to form a
bird's-eye view (BEV) feature map. This representation is then
processed by a conventional dense CNN. Finally, post-processing
operations, such as non-maximum suppression (NMS), are applied to
generate the final detection results.

\subsection{Convolution and Batch Normalization}
As discussed above, once the 3D point cloud input is encoded
into pillars or voxels, it can be processed by CNNs. Importantly, CNNs
do not require the spatial dimensions of their inputs (i.e., the
input resolution) to be fixed. Instead, these dimensions can be
determined at runtime, allowing the same CNN to process inputs of
varying resolutions.

It is worth noting that CNNs typically incorporate batch normalization (BN) layers following each convolution layer to accelerate and stabilize model training.
A BN layer is formally expressed as:
\begin{equation}
    y = \gamma \cdot \frac{x - \mu}{\sigma} + \beta
\end{equation}
where $y$ is the normalized output, $x$ is the input, $\mu$ and $\sigma$ denote the mean and standard deviation, respectively, and $\gamma$ and $\beta$ are learnable scale and shift parameters whose values depend on the statistical distribution of the input tensors~\cite{ioffe2015batch}.

\subsection{Resolution Scaling of Pillars/Voxels}
\label{background:varinputres}
% \yun{talk about 3D point cloud to pillars first, then talk about CNNs}
Given the general architecture of a LiDAR object detection DNN described above, adjusting the pillar or voxel size derived from point clouds represents an effective strategy for scaling the input resolution of the detector~\cite{pointnetpp,pillarnet}. Figure~\ref{fig:pillars} illustrates three examples of pillars generated from the same point cloud.
Enlarging the pillar dimensions $(V_x, V_y)$ used to encode the points reduces both the total number of pillars and the height and width of the grid $G$.
This leads to faster inference without requiring any modifications to the model architecture.
However, the trade-off is a reduced capacity to capture fine-grained spatial details, analogous to the loss of detail observed when processing lower-resolution images.

\section{Motivation}
\label{sec:motivation}

In this section, we explore the feasibility and challenges of resolution scaling to enable anytime computing capability in LiDAR object detection.

\begin{figure}[htp]
\centerline{\includegraphics[scale=0.56]{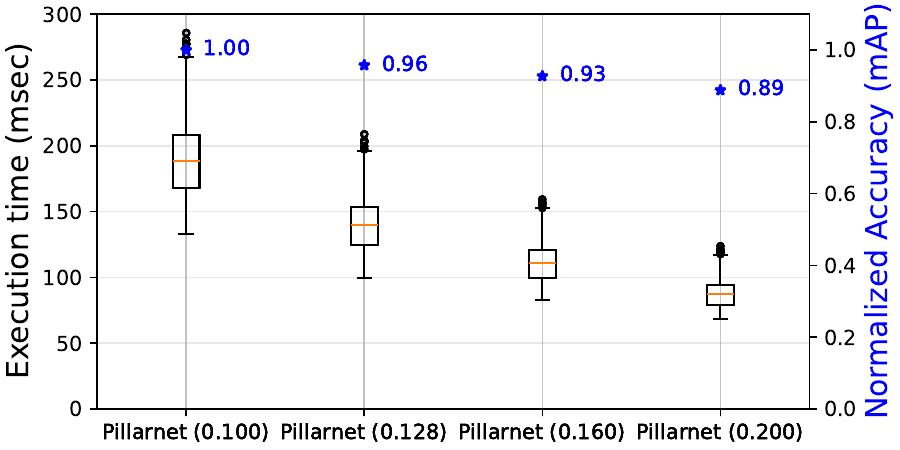}}
\caption{Execution time (on Jetson AGX Orin at 30W) and accuracy statistics of Pillarnet separately trained with four different pillar sizes.}
\label{fig:e2eboxplot}
\end{figure}

\begin{figure*}[t]
\centerline{\includegraphics[scale=0.65]{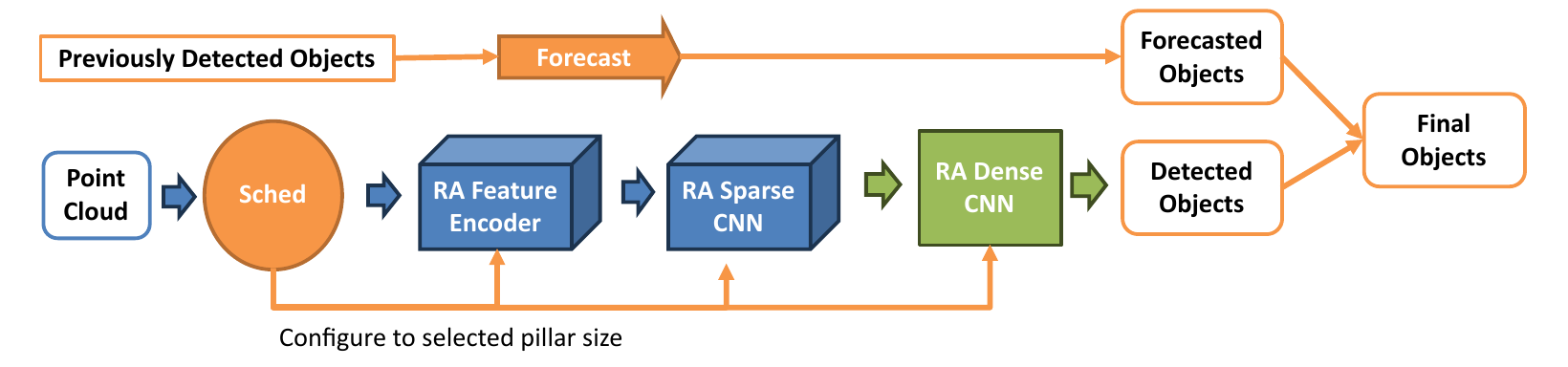}}
\caption{The architecture of MURAL. RA stands for resolution-aware.}
\label{fig:mural}
\end{figure*}

Figure~\ref{fig:e2eboxplot} shows the execution time distribution and average accuracy of Pillarnet~\cite{pillarnet} LiDAR object detection models, where each model is trained for a different input resolution. As expected, higher resolutions resulting from smaller pillar sizes enable higher accuracy but also, on average, longer execution times. A straightforward way to enable anytime computing would be to deploy multiple models at different resolutions and switch among them based on the current deadline. However, this strategy requires loading multiple DNNs into memory, which is often impractical on memory-constrained embedded platforms.

In principle, one could instead run a single model at resolutions other than the one for which it was trained, leveraging the fully convolutional nature of LiDAR object detection architectures. However, Table~\ref{tab:pillar_accuracy} shows the normalized mAP scores of Pillarnet when trained at $0.100^2\,\text{m}^2$ and evaluated with four different pillar sizes. As shown, when the resolution used during inference differs from the training resolution, accuracy drops significantly. Thus, naively using a single model for multiple resolutions is not a viable solution for anytime computing.

\begin{table}[htp]
    \centering
    \small
    \begin{tabular}{|l|cccc|}
        \hline
        \textbf{Pillar size ($m^2$)} & $0.100^2$ & $0.128^2$ & $0.160^2$ & $0.200^2$ \\ \hline
        \textbf{Normalized mAP (\%)} & 100.0 & 78.8 & 41.0 & 18.0 \\ \hline
    \end{tabular}
    \caption{Impact of pillar size mismatch.}
    \label{tab:pillar_accuracy}
\end{table}

In this paper, we propose a framework that equips a single LiDAR object detection model with anytime inference capability in a deployment-friendly manner. Specifically, our objective is to support multiple input resolutions at runtime while maintaining minimal memory overhead and preserving accuracy relative to baseline models trained at fixed resolutions.

\section{MURAL}
\label{sec:MURAL}

% \begin{figure*}[t]
% \centerline{\includegraphics[scale=0.40]{figures/bn_params_plot.pdf}}
% \caption{The architecture of MURAL.}
% \label{fig:bnparams}
% \end{figure*}

In this section, we introduce MURAL, a \textbf{MU}lti-\textbf{R}esolution \textbf{A}nytime \textbf{L}iDAR framework, which transforms any pillar- or voxel-based LiDAR object detection DNN into an (non-interruptible~\cite{zilberstein1999contract}) anytime algorithm, ensuring that detection results are delivered on time with the highest possible accuracy.

Throughout this section, we use "pillar" to refer to both pillar- and voxel-based representations for brevity. We provide explicit clarification for cases where the implementation or applicability differs between the two.

\subsection{Overview}
\label{sec:overview}

The general architecture of MURAL is illustrated in Figure~\ref{fig:mural}. MURAL is designed to enable efficient accuracy–latency trade-offs by dynamically adjusting the size of pillars into which the input point cloud is encoded.
Recall that increasing the pillar size reduces the number of pillars and the resolution (i.e., the width and height dimensions) of the grid processed by the CNN (Section~\ref{background:varinputres}).
By adjusting the pillar size, MURAL ensures that the entire input is processed within the deadline for each invocation of the object detector.
To support this capability, MURAL modifies the normalization layers of the target DNN to be resolution-aware and trains the model to adapt to multiple pillar sizes (Section~\ref{sec:multires_training}).
After training, MURAL can support additional pillar (only, not voxel) sizes—beyond those used during training by regressing resolution-specific parameters (Section~\ref{sec:posttrainingpillars}). 

During inference, its scheduler (Section~\ref{sec:dl_aware_sched}) takes the input point cloud and selects the smallest pillar size that can meet a given deadline, as smaller pillars are most likely to yield better detection performance. This selection entails accurately predicting the end-to-end execution time for each candidate pillar size until one that can meet the deadline is found. The scheduler then configures the detection pipeline to accommodate the chosen pillar size.

Beyond its core multi-resolution capabilities, MURAL integrates complementary optimizations from our prior work~\cite{VALO2024}. Specifically, it eliminates redundant computations in the dense CNN layers (Section~\ref{sec:dense_cnn_opt}) and utilizes a forecasting mechanism to predict object positions based on historical detections (Section~\ref{sec:forecasting}). After forecasting, detected and forecasted objects are merged, with priority given to the current detections to maximize accuracy. Furthermore, for voxel-based models, MURAL incorporates a region-dropping mechanism (Section~\ref{sec:reg_drop}) to mitigate the impact of potential execution time prediction inaccuracies.
Following the execution of the 3D backbone, this mechanism crops the intermediate data to reduce the execution time if a deadline miss is anticipated otherwise.

\begin{figure*}[t]
\centerline{\includegraphics[scale=0.6]{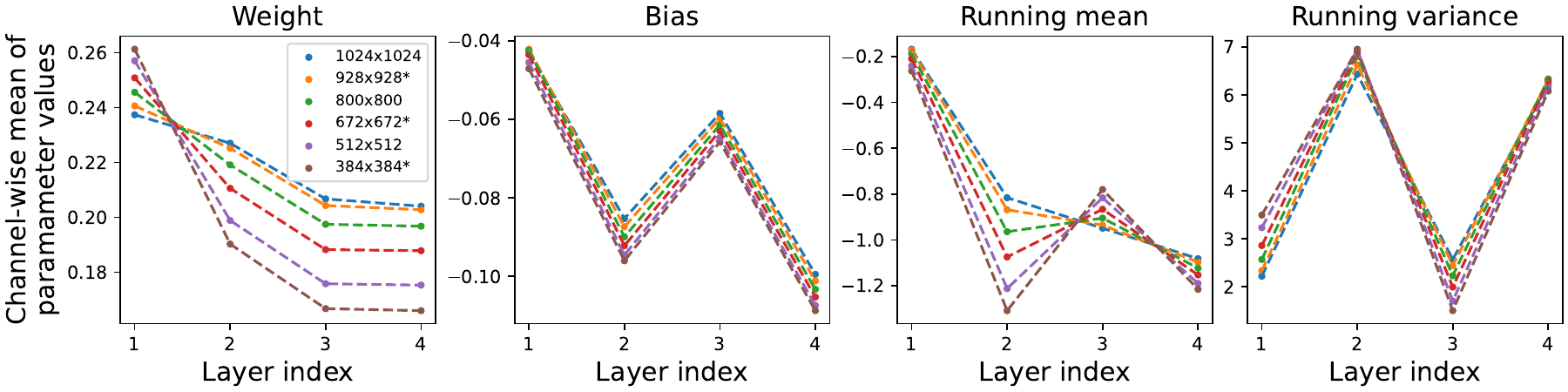}}
\caption{Channel-wise means of batch normalization parameters (weight, bias, running mean, running variance) for six grid areas from $1024 \times 1024$ to $384 \times 384$. The grid areas of the synthesized resolutions are indicated with a (*) in the legend. Their predicted batch normalization parameters are predicted with regression.}
\label{fig:bnparams}
\end{figure*}

\begin{figure}[htp]
\centerline{\includegraphics[scale=0.6]{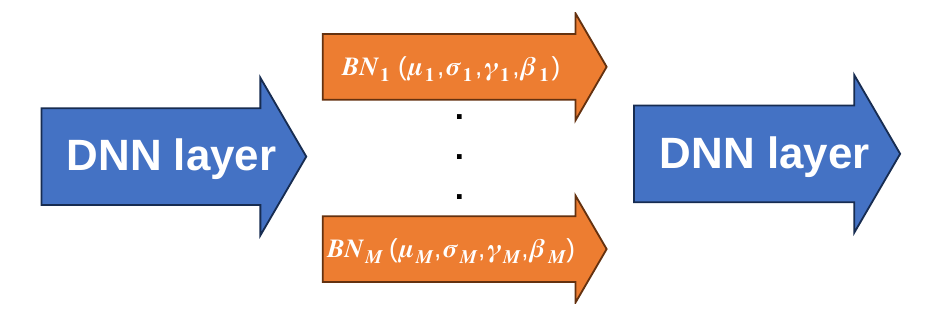}}
\caption{Resolution-aware batch normalization.}
\label{fig:resawarebn}
\end{figure}

\subsection{Multi-Resolution Training and Inference}
\label{sec:multires_training}
We introduce a DNN training scheme for LiDAR object detection that allows the pillar size to be dynamically selected at runtime from a predefined set of options.
A key requirement is that for each pillar size, the single trained model should achieve accuracy comparable to or exceeding that of individually trained dedicated models.
Consequently, a MURAL-enabled DNN can serve as a drop-in replacement for multiple baseline DNNs, facilitating efficient and memory-friendly accuracy-latency trade-offs.

In the proposed training scheme, for each input batch of point clouds, a separate forward pass is conducted for every targeted pillar size, and the resulting loss values are accumulated as follows:
\begin{equation}
    \mathcal{L}_{\text{total}} = \sum_{p \in \mathcal{P}} \mathcal{L}(f_\theta(x, p), y)
\end{equation}
where $\mathcal{L}_{\text{total}}$ is the total accumulated loss, $\mathcal{P}$ is the set of all targeted pillar sizes, $\mathcal{L}$ is the loss function of the baseline DNN, $f_\theta$ is the DNN with weights $\theta$, $x$ is the input point cloud encoded into pillars of size $p$, and $y$ is the ground truth.

Subsequently, backpropagation is applied using $\mathcal{L}_{\text{total}}$ to update the model parameters $\theta$, ensuring that the gradients accumulated across all resolutions contribute to each parameter update. While this approach renders the DNN adaptable to $\mathcal{P}$ to some extent, we observe that the accuracy achieved for each pillar size still falls noticeably below that of separately trained models.

To overcome this limitation, we introduce dedicated batch normalization (BN) layers for each input resolution throughout the DNN, as depicted for a single BN layer in Figure~\ref{fig:resawarebn}.
This design draws inspiration from a prior work on image classification~\cite{rsnets}, which demonstrates that varying image resolutions give rise to distinct statistical distributions that influence the behavior of BN layers (see Section~\ref{background:varinputres}).
That work proposes resolution-aware BN layers as an effective mechanism for accommodating multiple image resolutions while keeping the weights of all other layers shared.

We hypothesize that LiDAR pillar feature maps exhibit resolution-dependent statistical distributions analogous to those observed in camera-based object detection networks, which motivates our decision to replace all BN layers with resolution-aware counterparts in LiDAR object detection networks.

During training, since forward passes are performed for each pillar size, all BN layers within each resolution-aware BN block are activated. As a result, backpropagation updates the parameters of the BN layers corresponding to every input resolution.
At inference time, the appropriate BN layers are dynamically activated based on the selected pillar size.
It is worth noting that since each BN layer has only a small number of parameters, maintaining separate layers per resolution introduces negligible memory overhead while yielding substantial accuracy improvements.

\begin{figure*}[htp]
\centerline{\includegraphics[scale=0.6]{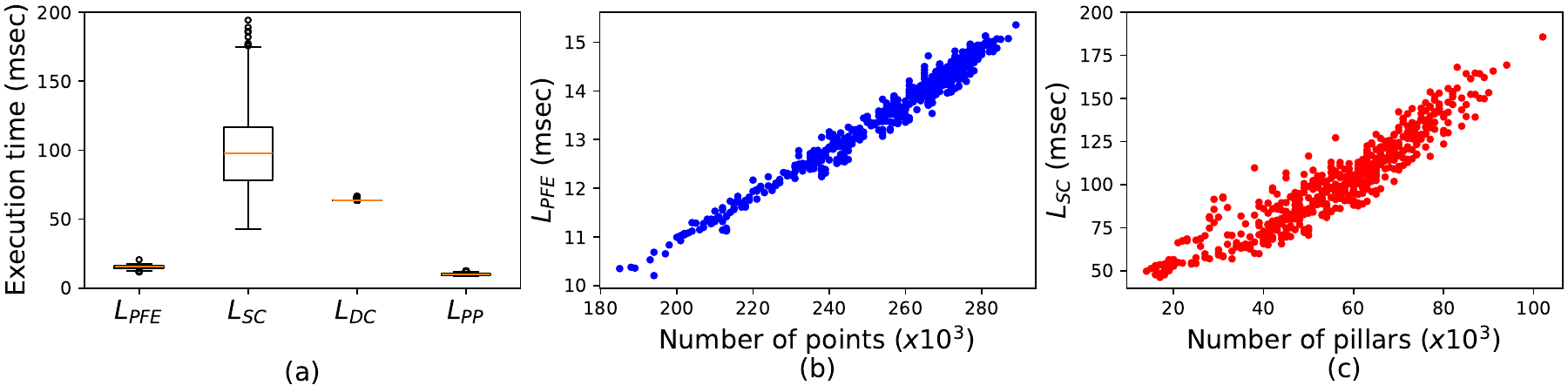}}
\caption{(a) Component-wise execution timing of the Pillarnet (trained for pillar size $0.100^2\,\text{m}^2$). (b) PFE latency of the same Pillarnet with respect to its input. (c) Sparse CNN latency of the same Pillarnet with respect to its input.}
\label{fig:pillarnet010_exec_time}
\end{figure*}

\begin{figure}[htp]
\centerline{\includegraphics[scale=0.45]{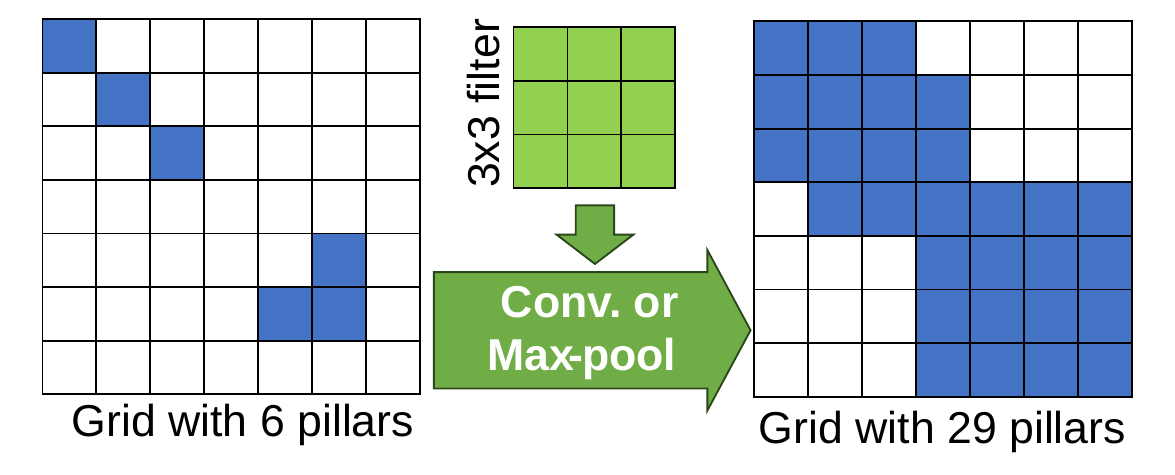}}
\caption{
Active output locations of a convolution depend on the spatial distribution of active input locations — a dependency that max pooling can replicate with matching kernel parameters.
}
\label{fig:conv_maxpool}
\end{figure}

\subsection{Synthesizing Additional Input Resolutions Post-Training}
\label{sec:posttrainingpillars}
% \yun{subsection title sounds strange. "pillar size" doesn't give a good intuition, while "resolution" does. why not keep the old title?}
At inference time, restricting the model only to input resolutions used during training can result in coarse-grained latency–accuracy trade-offs. To provide greater flexibility, we enable the use of additional resolutions at runtime by synthesizing BN layers post-training, thereby widening the list of available pillar sizes without requiring retraining or fine-tuning. The parameters of these synthesized BN layers are predicted using regression models fitted to the parameters of the originally trained BN layers. Specifically, for each BN parameter — running statistics (mean and variance) and learnable parameters (scale and shift) — we fit independent quadratic models per layer and input channel.

Figure~\ref{fig:bnparams} illustrates the BN parameters for six different input resolutions (resulted from six different pillar sizes). The channel-wise means are visualized to demonstrate that the relationship between the input resolutions and the parameter values can be modeled independently for each layer. Notably, only three of the input resolutions were used during training; the remaining three (marked with `*') were synthesized post-training.

This approach allows our model to generalize to resolutions beyond those encountered during training. Empirically, we observe that the accuracy achieved with additional resolutions falls between the accuracies of the two closest trained ones. Furthermore, adding resolutions smaller than the smallest trained resolution still yields satisfactory accuracy.

Unfortunately, the input resolution synthesis introduced here does not produce satisfactory results for voxel-based models. Although their BN layers also exhibit a strong relationship between input resolution and parameter values (similar to Figure~\ref{fig:bnparams}), the effectiveness of the synthesized BN layers also depends on the adaptability of the convolutional layers to the new resolutions. We posit that the sparse 3D CNNs used in voxel-based models struggle in this regard, as the 3D convolutions are more sensitive to resolution changes than the 2D convolutions used in pillar-based models.

\subsection{Dynamic Resolution Scheduling}
\label{sec:dl_aware_sched}

% \yao{Section 4.4 has a lot of moving parts. Would a small Algorithm 1 pseudocode block help to improve readability?}

To maximize detection accuracy under a dynamically imposed deadline, we propose scheduling the smallest pillar size (i.e., the highest input resolution) that can be processed within the given deadline.
This necessitates accurate runtime latency prediction for multiple pillar sizes, which is a non-trivial challenge due to the high variability in execution times.

To address the latency prediction challenge for a given pillar size, we decompose the total latency $L$ of a LiDAR object detection model into four constituent parts:
\begin{equation}
    L = L_{PFE} + L_{SC} + L_{DC} + L_{PP}
\end{equation}
where the four components represent the latencies of the pillar feature encoder (PFE), sparse CNN, dense CNN, and post-processing stages, respectively. Note that this decomposition equally applies to voxel-based models by substituting the PFE  with the voxel feature encoder (VFE).

Figure~\ref{fig:pillarnet010_exec_time}\textcolor{blue}{-a} shows the latencies of these four components for baseline Pillarnet~\cite{pillarnet} utilizing a pillar size of $0.100^2\,\text{m}^2$. We first observe that $L_{PP}$ and $L_{DC}$ show almost no variation in execution time across a range of input samples. Therefore, their 99th percentile values, obtained from offline benchmarking per input resolution, provide reliable predictions for all MURAL pillar size configurations.

For $L_{PFE}$, shown in Figure~\ref{fig:pillarnet010_exec_time}\textcolor{blue}{-b}, there is a strong correlation between the number of input points and $L_{PFE}$. For each input resolution, we model this relationship with a quadratic regression and use these fitted models for runtime prediction.

Finally, $L_{SC}$ exhibits substantial variability, as illustrated in Figure~\ref{fig:pillarnet010_exec_time}\textcolor{blue}{-c}, even when processing an identical number of input pillars. This unpredictability renders execution time estimation unreliable using either a fixed worst-case value or a simple quadratic equation as employed for $L_{PFE}$.

This variability arises because each sparse convolution within the sparse CNN can produce a different number of output pillars for the same number of input pillars. Since sparse convolution filters operate based on the spatial coordinates of active input pillars within the grid, both the count and coordinates of the output pillars are inherently dependent on those of the inputs. An illustrative example is provided in Figure~\ref{fig:conv_maxpool}. This spatial dependency introduces a cascade effect throughout the sparse CNN, causing the number of active pillars to fluctuate dynamically across layers---the primary driver of execution time variability~\cite{VALO2024}.

While the relationship between input pillar count and execution time for any individual sparse convolution can be accurately captured by a quadratic model, the core difficulty lies in predicting per-layer pillar counts prior to sparse CNN execution. Our previous work~\cite{VALO2024} addressed this through a history-based approach that assumed strong spatial consistency between consecutive LiDAR frames. However, this assumption weakens as the environment becomes increasingly dynamic.

In this work, we propose a more robust latency estimation method for pillar-based models that infers the input pillar counts for all sparse convolution layers without executing the sparse CNN itself. Our approach employs lightweight max pooling operations configured to replicate the behavior of the sparse convolution layers. The underlying insight is that for a given input, the output pillar coordinates of a convolution (sparse or dense) can be reproduced by a suitably configured max pooling operation sharing the same kernel size, stride, and padding as the convolution it emulates.
Once the input pillar counts for all sparse convolution layers are calculated via max pooling, $L_{SC}$ is predicted by mapping these counts to execution times through per-layer quadratic equations and summing the results.
These calculations are carried out independently for each pillar size considered during scheduling.

For voxel-based models, the overhead of 3D max pooling is prohibitive, rendering our max-pooling based time prediction infeasible. Unlike pillar-based models where sparse CNN operates over a compact 2D representation, voxel-based models process the input in 3D. Thus, mimicking the sparse CNN of voxel-based models requires 3D max pooling operations (whether sparse or dense), making its execution cost disproportionately high relative to the 2D pillar-based counterpart.

To address this, we construct an offline lookup table where each row corresponds to a representative input sample and each column to a candidate resolution, with cell values storing the measured execution time. Samples are selected to span the expected range of scene densities and spatial distributions encountered at runtime, ensuring the table captures latency variability across diverse conditions. 
Our method is built on two premises: (1) consecutive inputs tend to share similar spatial characteristics, resulting in comparable latencies, and (2) relative latency trends across resolutions remain consistent for inputs with similar spatial characteristics.
At runtime, we identify the column of our lookup table corresponding to the pillar size used in the last frame, find the row whose latency value in that column is closest to the last frame's execution time, and use the latencies in that row as our predicted execution times for all candidate resolutions. Empirically, this approach yields prediction accuracy similar to the history-based method described in~\cite{VALO2024}. However, it inherits the spatial consistency assumption from~\cite{VALO2024}, and the region-dropping mechanism described in Section~\ref{sec:reg_drop} is employed to mitigate the impact of potential prediction inaccuracies.
\begin{algorithm}

\caption{Dynamic Resolution Scheduling}
\begin{algorithmic}[1]
\REQUIRE Deadline $d$, \\
input point cloud $C$,\\
available pillar sizes $P = \{p_1, p_2, \ldots, p_n\}$ (sorted from smallest to largest)
\ENSURE Selected pillar size $p^*$

\FOR{$i = 1$ to $n$}    
    \STATE $t_i \gets predict\_exec\_time(C, p_i)$ 
    \IF{$t_i \leq d$}
        \STATE \textbf{return} $p_i$
    \ENDIF
\ENDFOR

\STATE \textbf{return} $p_n$

\end{algorithmic}
\label{algo:sched}
\end{algorithm}

Algorithm~\ref{algo:sched} summarizes the steps of our dynamic resolution scheduling.
The scheduler performs a top-down search, starting from the highest resolution and decreasing it until the predicted latency falls within the deadline. MURAL then configures the DNN to match the selected resolution.

% MURAL dynamically reconfigures the DNN to accommodate the selected pillar size by following the steps below:
% \begin{enumerate}
%     \item Feature encoder is configured to encode the points into pillars of selected size.
%     \item The normalization layers of the selected pillar size are activated in feature encoder, sparse CNN, and dense CNN.
%     \item Post-processing is informed of the resolution change to make processing of output tensors correct, since the output tensor sizes change with respect to the input.
% \end{enumerate}

\subsection{Dense CNN Optimizations}
\label{sec:dense_cnn_opt}

The point cloud from LiDAR may occupy a smaller area in the BEV than the cuboid space defined by its range.
As a result, large portions of the grid, especially at the edges, can be empty.
We crop these empty regions to speed up the dense CNN without sacrificing accuracy, as implemented in our prior work~\cite{VALO2024}.
Additionally, dense convolutions in detection head that infer object attributes (e.g., size, velocity) perform computations across the entire grid, without considering object locations.
To avoid the redundant computations introduced by this, we apply an optimization from~\cite{VALO2024} that limits the computation to regions where detected objects are located, reducing latency while maintaining accuracy.
This optimization is also incorporated into MURAL.

\subsection{Forecasting}
\label{sec:forecasting}
The forecasting process entails predicting the current positions of previously detected objects based on their estimated velocity and ego-vehicle localization data.
In our prior work~\cite{VALO2024}, input data scheduling was employed to navigate latency--accuracy trade-offs, where a portion of the input data was skipped and forecasting of past detection results was used to compensate for the missing information.
Interestingly, forecasting also proves beneficial within MURAL, despite the absence of any input skipping. Specifically, it enables objects that were missed in the current frame (e.g., due to occlusion) but detected in earlier frames to be continuously tracked.
For this reason, forecasting is integrated into the MURAL framework.

\subsection{Region Dropping}
\label{sec:reg_drop}
Our lookup table-based sparse CNN time prediction method for voxel-based models can inevitably yield over- and underpredictions, since it assumes perfect spatial consistency across consecutive LiDAR scans — an assumption inherited from our prior work~\cite{VALO2024}. Underpredicted timings of the sparse CNN can lead to deadline misses. To mitigate this issue, we adapt the region dropping method introduced in~\cite{VALO2024} to MURAL. This method, executed after the sparse CNN, trims the dense input tensor to the largest size that can still meet the deadline, and is only applied when an underprediction would otherwise cause a deadline miss. Specifically, regions are equally cropped from the left and right sides of the BEV input tensor.
This is because the ego vehicle and its immediate surroundings are centered in the BEV representation as illustrated in Figure~\ref{fig:pillars}, making the left and right regions less likely to contain safety-critical objects.
\section{Evaluation}
\label{sec:evaluation}

We extended OpenPCDet~\cite{openpcdet2020}, an open-source LiDAR object detection framework that supports state-of-the-art methods, to implement MURAL.
Our evaluation is two-fold: an open-loop study using the nuScenes~\cite{nuscenes} dataset to assess detection performance under hard deadlines, and a closed-loop study in a simulated environment to evaluate the impact of MURAL on autonomous driving safety and efficiency.

\subsection{Open-loop Evaluation}
\label{sec:openloopeval}
In our open-loop evaluation, we primarily use Pillarnet~\cite{pillarnet}, a leading pillar-based DNN.
To demonstrate MURAL’s general applicability, we also present results on two more methods: PointPillars~\cite{pointpillars}, and CenterPoint~\cite{centerpoint}.
Table~\ref{tab:model_arch} presents the architectural differences of the evaluated models.
Unlike Pillarnet and CenterPoint, PointPillars does not employ a sparse CNN, which simplifies its scheduling because latency prediction becomes more straightforward.

\begin{table}[h]
    \centering
    \begin{tabular}{|l|c|c|c|}
        \hline
        & Feature Encoding & Sparse CNN & Dense CNN \\ \hline
        Pillarnet & Pillar & \checkmark & \checkmark \\ \hline
        PointPillars & Pillar & $\times$ & \checkmark \\ \hline
        CenterPoint & Voxel & \checkmark & \checkmark \\ \hline
    \end{tabular}
    \caption{Architectural differences of evaluated models.}
    \label{tab:model_arch}
\end{table}

For comparison, MURAL is evaluated against our prior work, VALO~\cite{VALO2024}, a state-of-the-art anytime LiDAR object detection framework that achieves anytime capability through input data slicing and scheduling. VALO is applied to baseline models utilizing the smallest available pillar/voxel size to maximize detection accuracy.

For model training and evaluation, we employ the nuScenes~\cite{nuscenes} autonomous driving dataset, reporting detection accuracy using the mean average precision (mAP) metric.
All models are trained on the full nuScenes training split, which comprises 700 distinct scenes, each consisting of a 20-second LiDAR scan sequence captured at 50-millisecond intervals.
In both training and evaluation, the 10 most recent LiDAR scans are merged for each input, a commonly adopted technique to enhance accuracy and facilitate object velocity estimation~\cite{nuscenes}.

For runtime evaluation under hard deadline constraints, we utilize 75 scenes from the nuScenes validation split, processing all annotated frames in each sequence sequentially at 250-millisecond intervals.
This process is repeated under varying deadline constraints for each evaluated model.
Our runtime testing methodology maintains a buffer of the most recent successful detection outputs, which is updated whenever a method meets its deadline.
In the event of a missed deadline, the late output is discarded and the buffered results are used instead, effectively simulating job abortion.

Runtime evaluation is conducted on two hardware platforms: NVIDIA Jetson AGX Xavier and NVIDIA Jetson AGX Orin, both operating under a 30W power profile.
Platform specifications are summarized in Table~\ref{tab:platforms}.
On each device, six CPU cores and all available GPU resources are exclusively allocated to the method under evaluation.

\begin{table}[h]
    \centering
    \begin{tabular}{|c|c|c|}
        \hline
                 & Jetson AGX Xavier & Jetson AGX Orin \\ \hline
        CPU      & 8-core NVIDIA Carmel     & 12-core Arm Cortex-A78AE    \\ 
        RAM      & 16 GB             & 32 GB        \\
        OS       & Ubuntu 20.04      & Ubuntu 22.04 \\
        Software & Jetpack 5.1       & Jetpack 6.0  \\ \hline
    \end{tabular}
    \caption{Experiment platforms.}
    \label{tab:platforms}
\end{table}

Our evaluation results are organized into six subsections: (1) details of MURAL's training; (2) MURAL’s performance on Pillarnet; (3) MURAL’s performance on PointPillars; (4) MURAL’s performance on CenterPoint; (5) an ablation study of MURAL’s components; (6) an analysis of the scheduler’s time prediction errors; and (7) overhead analysis.
In each section (except for the first), we normalize the detection scores (mAP) of all the methods evaluated relative to the highest score obtained in that section.

\subsubsection{Training Results}
\label{sec:training_eval}

Using the nuScenes dataset, we first train three separate baseline Pillarnet models, each with a distinct pillar size, and subsequently compare MURAL's performance across resolutions against these baselines.
Note that in this experiment, forecasting is disabled for MURAL and no deadline violations are assumed for any method, in order to isolate the accuracy implications of MURAL.

Table~\ref{tab:training1} presents the results.
MURAL maintains comparable accuracy for the smallest ($0.100^2$) and largest ($0.200^2$) pillar sizes, while surpassing the baseline for the medium pillar size ($0.128^2$).
These results demonstrate that MURAL, as a single model capable of operating across multiple resolutions, achieves accuracy on par with or better than individually trained fixed-resolution models.
We attribute the accuracy gain at the medium resolution to the regularization effect inherent in multi-resolution training, consistent with observations reported in~\cite{rsnets} for multi-resolution image classification.

\begin{table}[htb]
    \centering
    \begin{tabular}{|c|c|c|}
        \hline
        \textbf{Pillar size ($m^2$)} & \textbf{Pillarnet} & \textbf{MURAL}  \\ \hline
        $0.100^2$ & 0.564  & 0.564 (+0.000)  \\
        $0.128^2$ & 0.537  & 0.560 (+0.023)  \\
        $0.200^2$ & 0.506  & 0.499 (-0.007)  \\ \hline
    \end{tabular}
    \caption{Accuracy in mAP of baseline Pillarnet and the MURAL-applied version. Numbers in parentheses indicate the differences with respect to the baseline.}
    \label{tab:training1}
\end{table}

In the next experiment, to evaluate the effectiveness of additional pillar size support described in Section~\ref{sec:posttrainingpillars}, we introduce several pillar sizes post-training and evaluate their mAP scores. 

Table~\ref{tab:mural_Pillarnet_posttraining} shows the results.
Note that the blue color represents the pillar sizes added post-training. The results show that for these pillar sizes, enabled by synthesized BN layers, the model achieves satisfactory accuracy, falling between the neighboring trained pillar sizes.
The smallest new pillar size ($0.263^2m^2$) further allows tight deadlines to be met.

\begin{table}[h]
    \centering
    \begin{tabular}{|c|c|c|}
        \hline
        \textbf{Pillar size ($m^2$)}  & \textbf{Grid area} & \textbf{mAP} \\
        \hline
        $0.100^2$ & $1024^2$ & 0.564 \\
        \color{blue}{$0.109^2$} & \color{blue}{$928^2$} & \color{blue}{0.568}  \\
        $0.128^2$ & $800^2$  & 0.560 \\
        \color{blue}{$0.151^2$} & \color{blue}{$672^2$} & \color{blue}{0.540}  \\
        $0.200^2$ & $512^2$ & 0.499 \\
        \color{blue}{$0.263^2$} & \color{blue}{$384^2$} & \color{blue}{0.390}  \\
        \hline
    \end{tabular}
    \caption{MURAL on Pillarnet with post-training introduced pillar sizes (blue).}
    \label{tab:mural_Pillarnet_posttraining}
\end{table}

\begin{figure*}[t]
    \centering
    % First row
    \begin{minipage}[t]{0.48\textwidth}
        \centering
        \includegraphics[width=\textwidth]{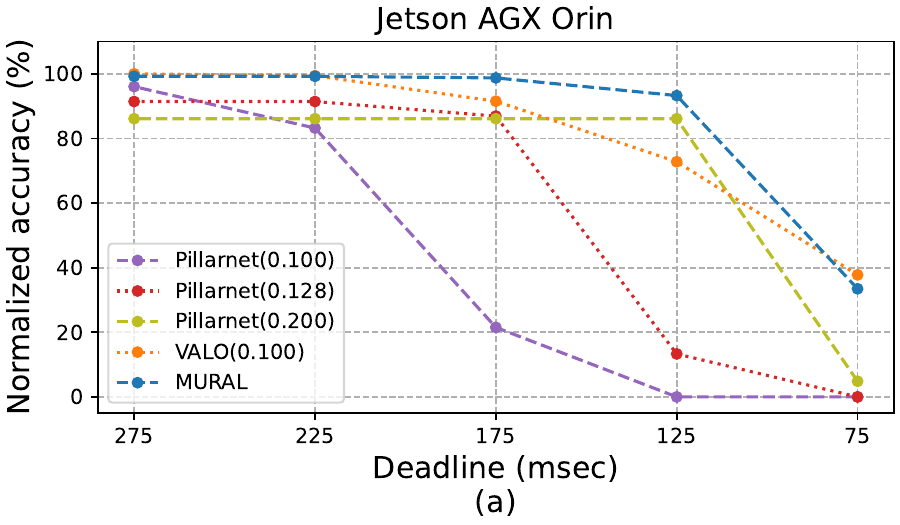}
        \label{fig:baseline_acc_orin}
    \end{minipage}%
    \hfill
    \begin{minipage}[t]{0.48\textwidth}
        \centering
        \includegraphics[width=\textwidth]{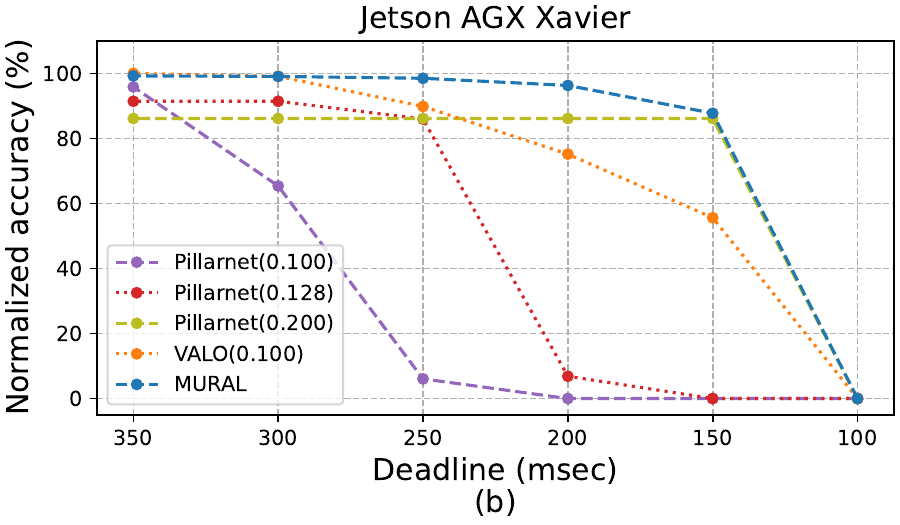}
        \label{fig:baseline_acc_xavier}
    \end{minipage}
    
    % \vspace{0.8cm}
    
    % Second row
    \begin{minipage}[t]{0.48\textwidth}
        \centering
        \includegraphics[width=\textwidth]{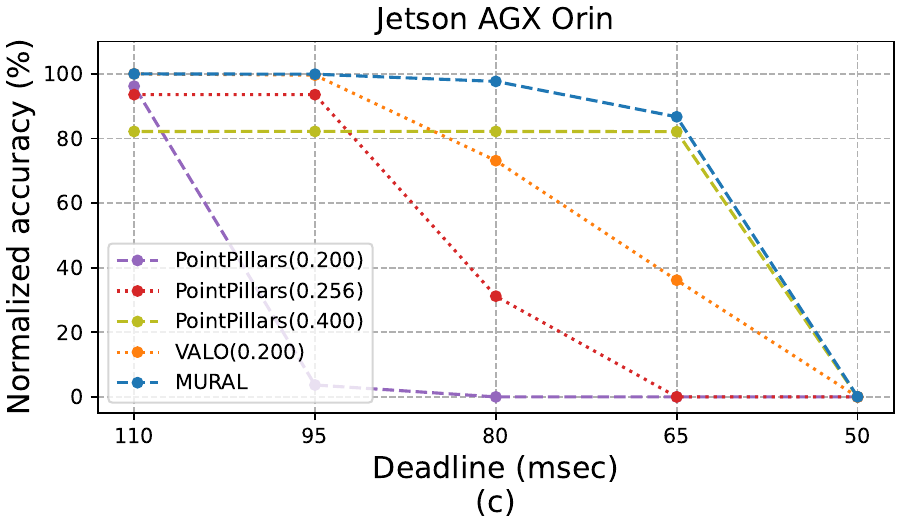}
        \label{fig:pp_acc_orin}
    \end{minipage}%
    \hfill
    \begin{minipage}[t]{0.48\textwidth}
        \centering
        \includegraphics[width=\textwidth]{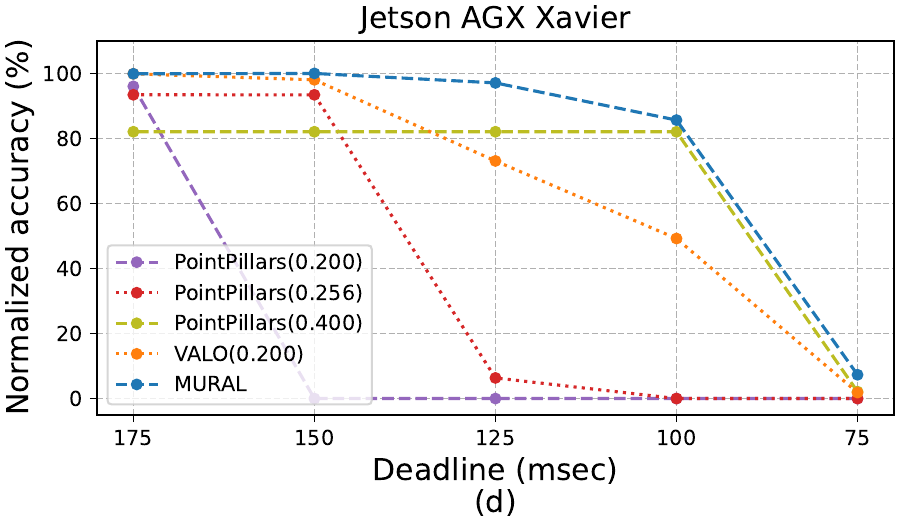}
        \label{fig:pp_acc_xavier}
    \end{minipage}

    % Third row - left and right aligned with 5% right-side trim
    \begin{minipage}[t]{0.48\textwidth}
        \centering
        \includegraphics[width=\textwidth, trim=5mm 0 7mm 0, clip]{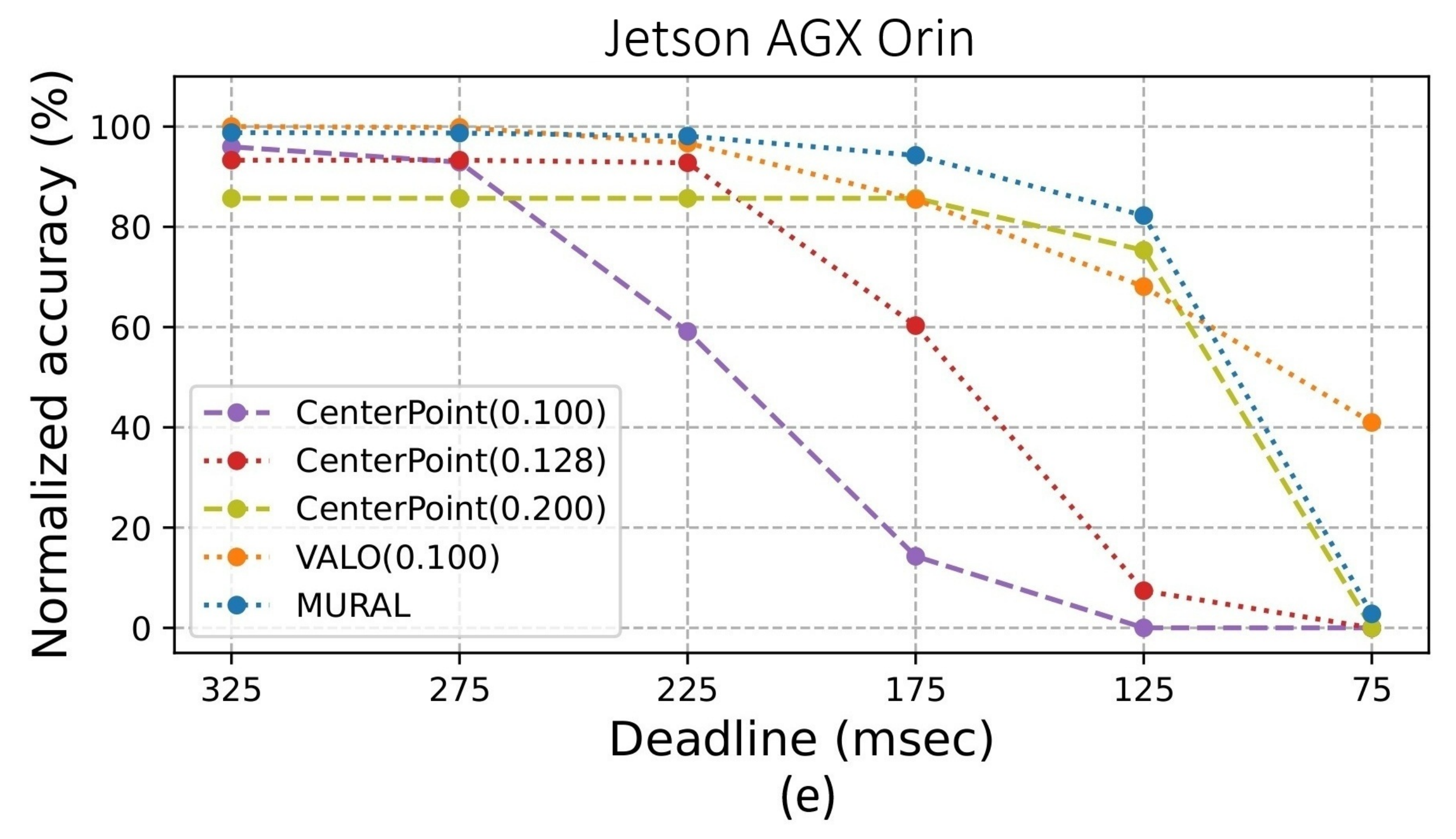}
        \label{fig:cp_acc_orin}
    \end{minipage}%
    \hfill
    \begin{minipage}[t]{0.48\textwidth}
        \centering
        \includegraphics[width=\textwidth, trim=5mm 0 7mm 0, clip]{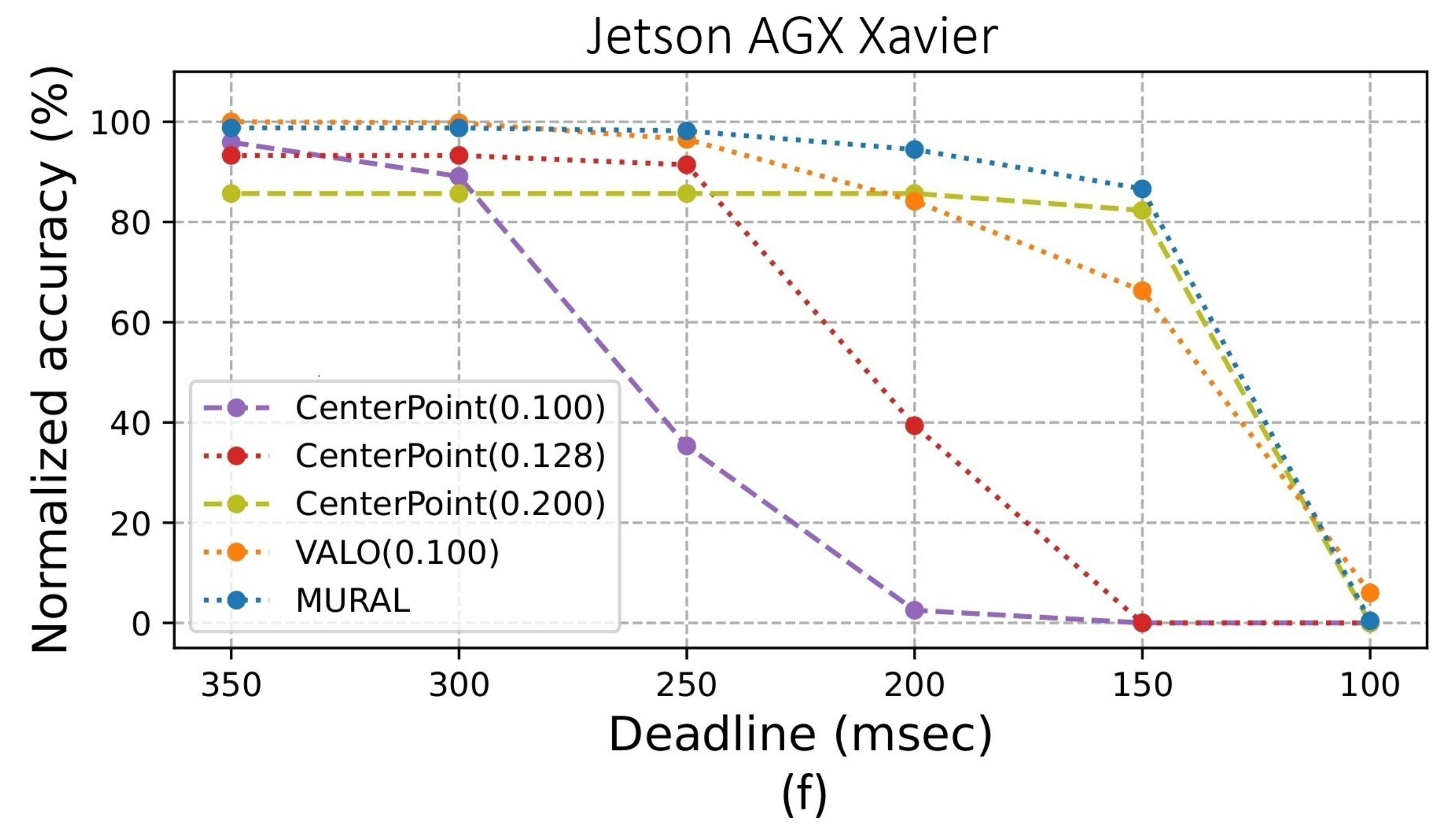}
        \label{fig:cp_acc_xavier}
    \end{minipage}
    \vspace{-10pt}
    \caption{(a,b) Pillarnet, (c,d) PointPillars, and (e,f) CenterPoint experiments on both evaluation platforms. In each plot figure, MURAL and VALO were applied to the baseline they are being compared.}
    \label{fig:combined_results}
    % \vspace{-10pt}
    \vspace{-1em}
\end{figure*}

\begin{figure}[t]
\centerline{\includegraphics[scale=0.57]{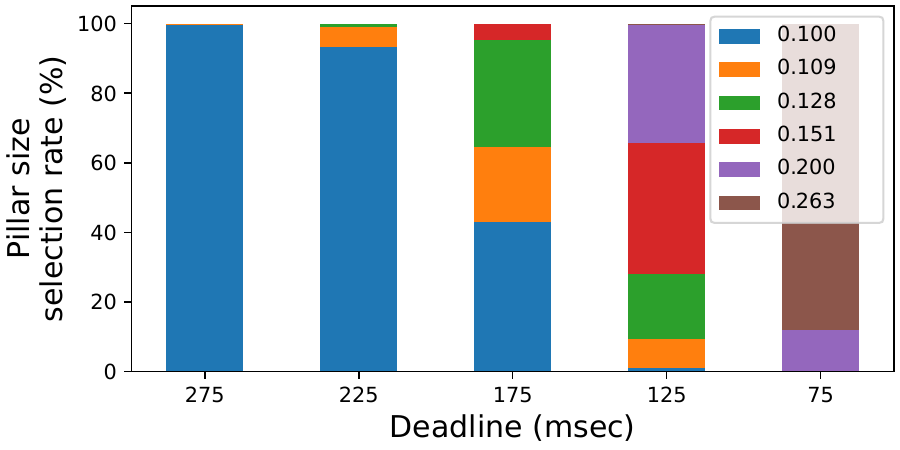}}
\caption{Pillar size selection rates of MURAL during Pillarnet experiment on Jetson AGX Orin.}
\label{fig:pillarrates}
\vspace{-2em}
\end{figure}

\subsubsection{MURAL on Pillarnet}
\label{sec:baseline_eval_Pillarnet}

Figures~\ref{fig:combined_results}\textcolor{blue}{-a} and~\ref{fig:combined_results}\textcolor{blue}{-b} depict the relationship between detection accuracy and deadline, with MURAL consistently outperforming the baselines on both platforms. By leveraging dynamic resolution scheduling, MURAL can draw from a broad range of resolutions to satisfy any given deadline, as shown in Figure~\ref{fig:pillarrates}, thereby maximizing accuracy under varying time constraints. The baseline Pillarnet models, lacking anytime computing capability, can only produce predictions within a much narrower deadline range, resulting in comparatively lower accuracies. VALO~\cite{VALO2024}, while possessing anytime capability and achieving higher accuracies than the baselines over a wider deadline range, is nonetheless outperformed by MURAL. Its data scheduling strategy proves less effective than MURAL's dynamic resolution scaling, primarily because VALO processes only a small fraction of the input data under tight deadlines, making it heavily reliant on forecasted detections. MURAL, by contrast, always processes the complete input frame---albeit at a reduced resolution---regardless of the imposed deadline. Consequently, MURAL achieves superior accuracy compared to both VALO and the individually trained baseline models.

\subsubsection{MURAL on PointPillars}
\label{sec:baseline_eval_pointpillars}

We further evaluate MURAL using the PointPillars~\cite{pointpillars} architecture, benchmarking it against multiple baseline PointPillars models as well as a VALO~\cite{VALO2024} variant applied to PointPillars. The MURAL model was trained to accommodate all pillar sizes used in the baselines, in addition to five supplementary synthesized pillar sizes introduced after training (see Section~\ref{sec:posttrainingpillars}).

Figures~\ref{fig:combined_results}\textcolor{blue}{-c} and~\ref{fig:combined_results}\textcolor{blue}{-d} present the results. Consistent with the Pillarnet findings, MURAL achieves better or comparable accuracy across all evaluated deadlines relative to both the PointPillars baselines and the VALO-applied model.

\subsubsection{MURAL on CenterPoint}
\label{sec:baseline_eval_centerpoint}

To showcase the performance of MURAL on voxel-based DNNs, we evaluate it using CenterPoint~\cite{centerpoint}.
For this task, we use multiple baseline CenterPoint models and VALO on CenterPoint.
The MURAL model was again trained to support all voxel sizes used in the evaluated baselines.

Figures~\ref{fig:combined_results}\textcolor{blue}{-e} and~\ref{fig:combined_results}\textcolor{blue}{-f} presents the results. Even without synthesizing additional pillar sizes (since this feature is only available for pillar-based models as explained in Section~\ref{sec:posttrainingpillars}), it still maintains better or comparable accuracy across all the tested deadlines in comparison to the baselines and VALO. The only exception is the tightest deadline on Orin, where MURAL's performance is noticeably worse than VALO's, due to MURAL having difficulty meeting deadlines with its smallest input resolution, which VALO can meet by processing only a very small subset of the input data. Nevertheless, the accuracy of VALO at this deadline is very low, and MURAL still outperforms the baselines by a large margin.

Overall, the results demonstrate that MURAL generalizes to multiple DNN architectures, both pillar- and voxel-based, and achieves efficient performance across different computing platforms, establishing it as the new state-of-the-art anytime LiDAR object detection method.

\subsubsection{Ablation study}
\label{sec:ablation_study}

We ablate MURAL on Pillarnet across four configurations: \textbf{SS} (static scheduler using WCET-based resolution selection), \textbf{DS} (dynamic scheduling of MURAL only), \textbf{DS-APS} (+ additional pillar sizes), \textbf{DS-APS-DCO} (+ dense CNN optimizations), and \textbf{DS-APS-DCO-FRC} (full MURAL with forecasting). No deadlines were missed in any configuration.

Table~\ref{tab:ablation} shows the results.
Using DS improves performance over SS due to its more accurate execution time prediction, enabling better scheduling decisions.
Introducing APS allows better utilization of the time until the deadline, improving accuracy.
Adding DCO allows meeting deadlines with higher resolutions without sacrificing accuracy. 
Finally, FRC further improves detection by recovering occluded or missed objects from prior frames.

\begin{table}[h]
\centering
\begin{tabular}{|l|c|c|c|}
\hline
\multirow{2}{*}{MURAL variant} & \multicolumn{3}{c|}{Deadline (ms)} \\
\cline{2-4}
 & 225 & 175 & 125 \\
\hline
SS & 96.17 & 85.60 & 85.60 \\
DS & 96.70 & 95.08 & 87.68 \\
DS-APS & 97.03 & 96.14 & 89.66 \\
DS-APS-DCO & 96.93 & 96.46 & 91.28 \\
DS-APS-DCO-FRC & 100.00 & 99.49 & 93.98 \\
\hline
\end{tabular}
\caption{Normalized accuracy of MURAL variants.}
\label{tab:ablation}
\end{table}

We also train MURAL without resolution-aware BN and make a comparison in Table~\ref{tab:ablation2}. Resolution-aware BN significantly improves performance, while using common BN layers for all pillar sizes degrades accuracy and makes it mostly adapt to the medium pillar size.

\begin{table}[h]
    \centering
    \begin{tabular}{|c|c|c|}
        \hline
        \textbf{Pillar size ($m^2$)} & \textbf{MURAL w/o RABN} & \textbf{MURAL w/ RABN}  \\ \hline
        $0.100^2$ & 75.00 & 100.00 \\
        $0.128^2$ & 93.373 & 99.238 \\
        $0.200^2$ & 61.836 & 88.395 \\ \hline
    \end{tabular}
    \caption{Normalized accuracy of MURAL (on Pillarnet) without and with resolution-aware BN. Forecasting was disabled and no deadline was considered.}
    \label{tab:ablation2}
\end{table}

\begin{figure}[t]
\centerline{\includegraphics[scale=0.57]{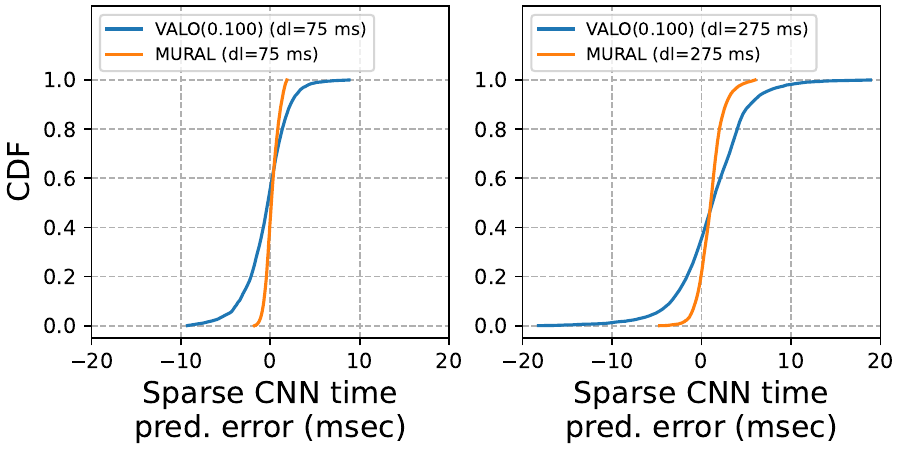}}
\caption{Time prediction errors of VALO and MURAL applied Pillarnet on Jetson AGX Orin. The errors were calculated by subtracting the actual time from the predicted time.}
\label{fig:timeprederr}
\end{figure}

\subsubsection{Time Prediction Error}
\label{sec:tprederr}

In this experiment, we examine the latency prediction error of MURAL's scheduler and compare it against that of VALO~\cite{VALO2024}. The analysis focuses specifically on the sparse CNN of Pillarnet, as latency prediction for the dense CNN and post-processing stages is straightforward and handled identically in both VALO and MURAL.

Figure~\ref{fig:timeprederr} presents the latency prediction errors as cumulative distribution functions.
As shown, MURAL achieves superior prediction accuracy by precisely estimating the number of input pillars for each sparse convolution layer within the sparse CNN, accomplished by efficiently emulating all sparse convolutions through max-pooling operations.

VALO, by contrast, assumes that the per-layer input pillar counts remain consistent with the most recently observed historical values.
This assumption breaks down in highly dynamic environments, where the 3D structure of successive LiDAR scans can vary considerably. Consequently, VALO's history-based prediction approach yields higher errors compared to MURAL's method.

\subsubsection{Time and Memory Overhead Analysis}
\label{sec:overhead}

Table~\ref{tab:sched_overhead} shows the average scheduling overhead of MURAL variants, measured on the Jetson AGX Orin.
Recall that MURAL on CenterPoint incorporates region drop (RD), explained in Section~\ref{sec:reg_drop}, and does not support synthesizing additional pillar sizes (APS), as this feature is only applicable to pillar-based models (Section~\ref{sec:posttrainingpillars}).

For the static scheduler (SS), the overhead is negligible across all models since we simply use the WCETs acquired from offline benchmarking. Using the dynamic scheduler (DS) notably increases the overhead for Pillarnet, as sparse convolutions must be mimicked for time prediction. However, the overhead increases only slightly for PointPillars, which does not use sparse CNNs. For CenterPoint, the DS overhead is moderate, falling between Pillarnet and PointPillars, as it does time prediction of sparse CNN with a history-based method (Section~\ref{sec:dl_aware_sched}).

When we introduce post-training pillar size synthesis (DS-APS) for the pillar-based models, the scheduling overhead increases due to new pillar sizes considered for scheduling. Adding dense convolution optimization (DS-APS-DCO and DS-DCO) requires determining the empty parts of the input scene in bird-eye view, which increases scheduling overhead. However, cropping these empty parts accelerates the dense CNN, compensating for the overhead. 

Enabling forecasting (DS-APS-DCO-FRC and DS-DCO-FRC) incurs no significant overhead, as it occurs in parallel on the CPU while the DNN layers execute on the GPU. Finally, adding region drop for CenterPoint (DS-DCO-FRC-RD) maintains nearly the same overhead, as the calculation of remaining time to deadline is minimal.

\begin{table}[t]
\centering

\begin{tabular}{|l|c|c|c|}
\hline
\multirow{2}{*}{MURAL variant} & \multicolumn{3}{c|}{Applied baseline} \\
\cline{2-4}
 & Pillarnet & PointPillars & CenterPoint \\
\hline
SS & 0.31 & 0.13 & 0.42 \\
DS & 3.23 & 0.53 & 1.19 \\
DS-APS & 5.47 & 1.17 & -- \\
DS-APS-DCO & 6.22 & 2.01 & -- \\
DS-APS-DCO-FRC & 6.24 & 1.97 & -- \\
DS-DCO & -- & -- & 1.76 \\
DS-DCO-FRC & -- & -- & 1.77 \\
DS-DCO-FRC-RD & -- & -- & 1.75 \\
\hline
\end{tabular}

\caption{Average scheduling overhead (milliseconds) of MURAL variants on Jetson AGX Orin.}
\label{tab:sched_overhead}
\end{table}

Finally, Table~\ref{tab:storage} shows the memory overhead of MURAL compared to using multiple baseline models with different resolutions. Note that MURAL, despite supporting multiple input resolutions, uses almost the same number of parameters as a single baseline model that supports only one resolution. This is because MURAL's memory overhead for supporting a new resolution is limited to the parameters for added BN layers, which are minimal compared to all the weights of the DNN. As a result, MURAL's memory overhead increases only slightly as a function of the number of resolutions it supports, whereas the memory overhead of the baseline models increases multiplicatively with the number of supported resolutions.

\begin{table}[h]
\centering
\begin{tabular}{|c|c|c|c|}
\hline
& Pillarnet & PointPillars & CenterPoint \\
% \cline{2-3}
 % & Pillarnet & PointPillars \\
\hline
Baseline & $61.003 \times 6$ & $23.956 \times 6$ & $35.766 \times 3$\\
MURAL    & 61.378 & 24.259 & 35.871 \\
\hline
\end{tabular}
\caption{Memory in MiB (megabytes) needed to store DNN parameters in 32-bit floating-point format.}
\label{tab:storage}
\vspace{-1em}
\end{table}

\subsection{Closed-loop evaluation}
\label{sec:closedloopeval}

To evaluate the impact of MURAL on navigation performance, we integrate it into the Autoware~\cite{autoware}, an open-source autonomous driving framework, and conduct closed-loop simulations using AWSIM~\cite{awsim}, a 3D game engine-based simulator.
Videos of our experiments are available~\footnote{Video link: \url{https://www.youtube.com/watch?v=Xi1YL2ukNHI}}.

\subsubsection{Simulation Setup}

\begin{figure}[htp]
\centerline{\includegraphics[scale=0.45]{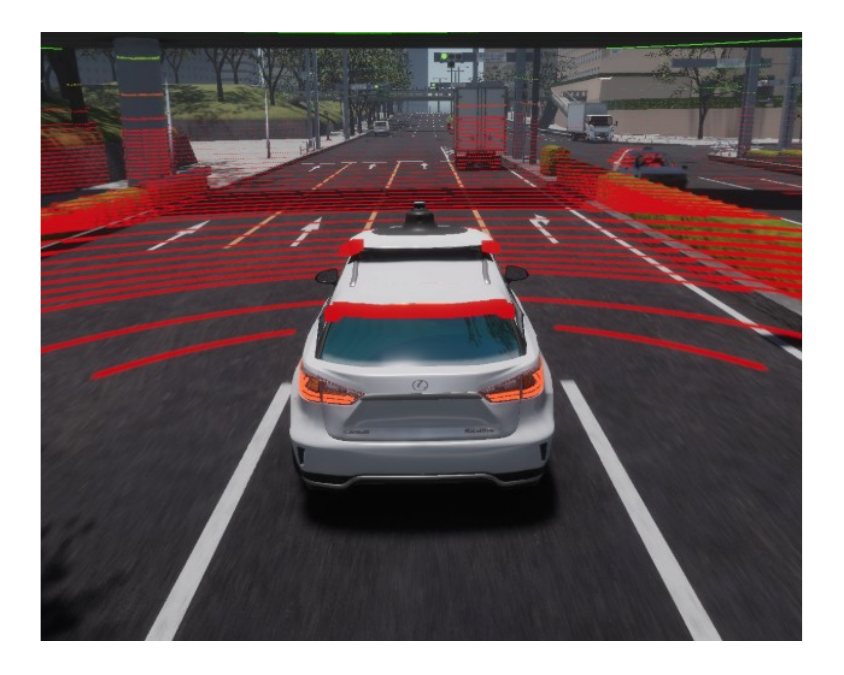}}
\caption{AWSIM simulator. Red points visualize a sample LiDAR scan.}
\label{fig:awsim}
\end{figure}

The simulation forms a closed-loop system where AWSIM (running at 100 frames per second) provides LiDAR point clouds and other sensor data to Autoware, while Autoware sends control commands to the simulator.
These data and command exchanges occur asynchronously.
Figure~\ref{fig:awsim} illustrates the environment simulated by AWSIM. 

We utilize Autoware's LiDAR-centric perception pipeline, where object detection results are processed by a Kalman filter-based tracker, then forwarded to a motion prediction task.
The predicted object trajectories are passed on to multiple planning tasks as the final result of object recognition.
To isolate the impact of the object detection DNN on driving, we disable the auxiliary LiDAR point cloud clustering task and ground-filter-based emergency stops.

We conduct the experiments on a desktop PC (AMD 7800X3D, NVIDIA RTX 4090, 128 GB DRAM, Ubuntu 22.04).
Since this hardware significantly outperforms embedded platforms, we emulate the execution time of a Jetson AGX Orin (30W mode) for the object detection DNN task using the latency prediction models described in Section~\ref{sec:dl_aware_sched}.

\subsubsection{Methodology}

We compare MURAL applied on Pillarnet against three fixed-resolution Pillarnet baselines ($0.100^2 \,\text{m}^2$, $0.128^2\,\text{m}^2$, and $0.200^2\,\text{m}^2$).
MURAL is configured to dynamically support these three resolutions and also synthesized ones (Section~\ref{sec:posttrainingpillars}).

We determine the object detection task deadline \textit{d} (milliseconds) at runtime based on the ego-vehicle velocity \textit{v} (m/s) using a linear mapping of $d = 250 - 12.5v$.
% At standstill (v = 0 m/s), the deadline is 250 ms, permitting the highest-resolution detection.
% At maximum speed (v ≈ 13 m/s), the deadline reduces to approximately 87.5 ms, enforcing low-latency detection to ensure timely reactions.
This mapping prioritizes high-resolution (thus, high-latency) detection at low speeds and low-latency detection at high speeds~\cite{d3}. We assume the maximum speed of the ego-vehicle to be around 13 m/s.

We evaluate performance using two distinct driving scenarios in an urban environment. (1) In the Dense Urban Environment, we measure the time to reach the destination in a cluttered setting where the ego-vehicle's path is clear but surrounded by numerous static and moving objects. (2) In the Dynamic Hazard Environment, we count collisions across three high-speed reactive scenarios: a suddenly appearing truck, a car merging into the ego-vehicle's lane, and a jaywalking pedestrian. We repeat both experiments 10 times and present the statistical results.

\subsubsection{Results in the Dense Urban Environment}

We evaluate a traffic intersection scenario featuring numerous pedestrians and vehicles and present our results in Table~\ref{tab:sim_results_dense}. Although the ego-vehicle's planned path was clear of obstacles, the low-resolution baseline ($0.200^2\text{m}^2$) frequently misclassified static objects, such as lampposts, as pedestrians. At lower resolutions, fine-grained geometric features of thin vertical structures are lost during voxelization, causing them to resemble the sparse point cloud signature of a pedestrian. These false positives (FPs) led the system to anticipate imminent crosswalk entries, causing the vehicle to stall or decelerate unnecessarily. The intermediate resolution baseline ($0.128^2\text{m}^2$) also suffered from elevated FP rates and longer completion times, suggesting that the misclassification issue is not exclusive to the lowest resolution. The strong correlation between FP rate and completion time across all baselines confirms that false positive detections are the primary bottleneck in this scenario. Consequently, the median time to reach the destination increased by up to 59\% for the lowest resolution compared to the $0.100^2\text{m}^2$ baseline. By selecting the highest resolution permitted by the deadline — which tends to be finer at lower speeds — MURAL achieved a completion time comparable to the high-resolution baseline.
MURAL's FP rate exceeds that of the Pillarnet $0.100^2\text{m}^2$ baseline because it dynamically switches to larger pillar sizes at higher speeds to meet tighter deadlines, increasing FP rate during those frames.

\subsubsection{Results in the Dynamic Hazard Environment}

We evaluate three high-speed scenarios (approximately 8 m/s): a sudden truck appearance, a merging car, and a jaywalking pedestrian. Table~\ref{tab:sim_results_dynamic} presents the results. The high-resolution baseline ($0.100^2\text{m}^2$) failed to avoid 16 out of 30 hazards due to an average detection latency of 247.8 ms. The intermediate resolution baseline ($0.128^2\text{m}^2$) reduced collisions to 11/30 by lowering average latency to 172.2 ms, yet still failed to avoid all collisions, suggesting the latency reduction remains insufficient for high speeds. The low-resolution baseline ($0.200^2\text{m}^2$) avoided all collisions with a 99.9 ms average latency.
MURAL also avoided all collisions, while maintaining an average latency of 174.4 ms — comparable to the $0.128^2\text{m}^2$ baseline. This highlights the key advantage of dynamic resolution selection: by reducing latency precisely during high-speed frames where the deadline is tightest, MURAL meets timing requirements when it matters most, unlike fixed-resolution baselines, which cannot adapt to varying deadline demands.

\begin{table}[t]
    \centering
    \caption{Closed-loop simulation results for the Dense Urban Environment. We report median track completion time and false positive detection rate.}
    \label{tab:sim_results_dense}
    \begin{tabular}{lcc}
        \toprule
               & Median   & Object Detection \\
        Method & Time (s) & FP Rate (\%) \\
        \midrule
        Pillarnet $0.100^2\,\text{m}^2$ & 31.70 & 13.6  \\
        Pillarnet $0.128^2\,\text{m}^2$ & 48.05 & 21.5  \\
        Pillarnet $0.200^2\,\text{m}^2$ & 50.40 & 27.7  \\
        MURAL (Ours) & 36.10 & 20.2  \\
        \bottomrule
    \end{tabular}
\end{table}

\begin{table}[t]
    \centering
    \caption{Closed-loop simulation for the Dynamic Hazard Environment. We report collision counts out of 30 (10 runs per scenario × 3 hazardous objects per run) and average latency.}
    \label{tab:sim_results_dynamic}
    \begin{tabular}{lcc}
        \toprule
        Method & Collisions & Average Latency (ms) \\
        \midrule
        Pillarnet $0.100^2\,\text{m}^2$ & 16/30 & 247.8 \\
        Pillarnet $0.128^2\,\text{m}^2$ & 11/30 & 172.2 \\
        Pillarnet $0.200^2\,\text{m}^2$ & 0/30 & 99.9 \\
        MURAL (Ours)                    & 0/30 & 174.4 \\
        \bottomrule
    \end{tabular}
    \vspace{-1em}
\end{table}

\subsubsection{Discussion of Closed-loop Simulation Results}

The results demonstrate that no single fixed resolution is sufficient for all driving conditions. High resolution is essential in complex urban scenes to minimize false positives and avoid unnecessary stalls; conversely, low latency is critical at high speeds to ensure timely hazard detection. Notably, while the low-resolution baseline avoids collisions in high-speed scenarios, its elevated false positive rate makes it unsuitable for dense urban environments. By dynamically adapting its input resolution, MURAL provides the versatility required for both efficient and safe autonomous navigation across diverse conditions.
\section{Related Work}
\label{sec:related}

Cyber-physical systems require timely execution to ensure safety and operational efficiency. Conventional methods that rely on fixed deadlines determined at design time~\cite{autoware, apollo} struggle to adapt to dynamic execution time requirements~\cite{alcon2020timing,d3}.

\textbf{Anytime Deep Neural Networks:}
In recent years, "anytime" processing has emerged as a promising approach for perception DNNs, enabling flexible trade-offs between accuracy and latency to meet varying deadline constraints.
Lee et al.\cite{lee2020subflow} reduced computation time by selectively deactivating non-critical neurons while prioritizing essential ones. Kim et al.\cite{kim2020anytimenet} introduced incremental layer processing with early exit mechanisms for image classification networks. Yao et al.\cite{yao2020imprecisecomp} and Bateni et al.\cite{bateni2018apnet} investigated scheduling strategies for multiple DNN tasks through imprecise computation, employing early exits and layer-wise approximation techniques, respectively.
However, these approaches primarily target image classification, which differs substantially from the complexity inherent in object detection tasks.

\textbf{Anytime End-to-end DNNs:}
Chen et al.\cite{chen2025timelynet} proposed TimelyNet, which integrates a supernet into the image encoder of end-to-end autonomous driving pipelines. It dynamically samples subnets of this supernet at runtime to meet deadlines determined by the vehicle's velocity and acceleration. Their evaluation included closed-loop CARLA~\cite{Dosovitskiy17CARLA} simulations across multi-modal pipelines, demonstrating that this adaptive approach significantly improves driving safety and reduces collisions compared to static models. While TimelyNet is designed for end-to-end DNNs, MURAL's target is the industry-standard modular approaches~\cite{autoware,apollo} where object detection is a separate DNN task.

\textbf{Anytime Object Detection:}
Several works have explored deadline-aware object detection strategies. Kuhse et al.\cite{kuhse2025anytimeyolo} examined early exit mechanisms for YOLO architectures. Heo et al.\cite{heo2022multipath} designed a multipath network architecture to enable anytime perception capabilities. Hu et al.\cite{hu2021exploring} proposed adaptive resolution reduction in scene regions deemed less critical. Liu et al.\cite{liu2022rttasksched, liu2020removing} partitioned image frames into sub-regions based on criticality, leveraging LiDAR data to prioritize processing of important areas. Kang et al.\cite{kang2022dnnsam} applied a split-and-merge strategy, processing critical regions at high resolution while handling non-critical areas at lower resolution. Gog et al.\cite{pylot} proposed dynamically switching between different DNNs based on runtime conditions.

\textbf{Multi-Resolution Processing:}
Heo et al.\cite{heo2022rtscale} presented an adaptive image scaling approach that adjusts resolution based on the operational environment, training a unified DNN capable of multi-resolution inference. However, their evaluation did not include comparisons against single-resolution baseline models. Wang et al.\cite{rsnets} employed resolution-sensitive batch normalization and ensemble distillation for image classification. Zhu et al.\cite{zhu2021dynamicresolutionnetwork} integrated a resolution predictor network into their framework. Chin et al.\cite{adascale2019chin} developed a resolution predictor for video object detection by exploiting temporal consistency across frames. Notably, most prior work in anytime object detection and multi-resolution processing concentrates on 2D vision tasks and does not address the distinct characteristics of 3D LiDAR point cloud object detection.

\textbf{LiDAR Object Detection:}
LiDAR-based object detection plays a vital role in autonomous driving systems~\cite{lidaro}. With the availability of large-scale datasets~\cite{nuscenes,sun2020waymo}, research has primarily focused on enhancing detection accuracy and reducing inference latency~\cite{pointpillars, pillarnet, shi2020parta2, voxelnext, centerpoint, ada3d, liu2022spatial, li2023pillarnext}. While these models achieve strong performance on high-end computing platforms, their deployment on embedded and edge devices remains challenging due to stringent size, weight, and power (SWaP) constraints and limited computational resources.

\textbf{Anytime LiDAR Object Detection:}
Soyyigit et al.\cite{alidar} introduced Anytime-LiDAR, which enables anytime processing through early exits and detection head scheduling for LiDAR object detection DNNs not having a sparse CNN~\cite{pointpillars}. However, this approach shows limited effectiveness with models of higher accuracy where sparse CNN is essential\cite{centerpoint, voxelnext, pillarnet}. VALO~\cite{VALO2024} subsequently proposed a data-scheduling strategy that maximizes input processing within deadline constraints and forecasts skipped data to maintain accuracy. While VALO offers flexibility, it suffers from accuracy degradation under tight deadlines due to incomplete input processing. Yuhang et al.~\cite{yuhang2024flex} explored multi-modal BEV detection with dynamic skipping of camera processing and LiDAR scans. However, their data scheduling approach is not directly applicable to single-modality models, which is the focus of our work.

\textbf{Our Approach:}
While reducing input resolution can substantially decrease latency with minimal accuracy loss, achieving dynamic resolution adjustment within a single DNN that maintains or exceeds the accuracy of single-resolution baselines remains challenging. Unlike prior work on resolution adaptation for image classification and 2D object detection, our research addresses dynamic resolution inference specifically for real-time LiDAR-based 3D object detection, filling a critical gap in the literature.
Importantly, unlike most prior work that relies solely on offline metrics, we conduct closed-loop simulations to evaluate the impact of our anytime framework on autonomous driving safety and efficiency in dynamic environments.
% Importantly, our work also demonstrates the practical implications of our approach in a closed-loop simulation setting.
\section{Conclusion}
\label{sec:conclusion}

This paper presented MURAL, a multi-resolution anytime framework for LiDAR 3D object detection that balances detection accuracy and processing latency through dynamic resolution scaling. Our approach combines multi-resolution training with shared weights, resolution-aware batch normalization, post-training pillar size synthesis, and deadline-aware scheduling, providing a memory-efficient solution that eliminates the need to deploy multiple model variants.

We demonstrated MURAL's versatility across both pillar-based (Pillarnet, PointPillars) and voxel-based (CenterPoint) architectures. Open-loop experiments on the nuScenes dataset show that MURAL achieves higher detection accuracy across various deadlines compared to baseline models and the prior state-of-the-art anytime approach. Closed-loop simulation experiments validate MURAL's practical benefits, demonstrating collision-free navigation while avoiding unnecessary stalls in dynamic environments. These results establish MURAL as the state-of-the-art solution for real-time LiDAR object detection on resource-constrained embedded platforms.
\section*{Acknowledgments}\label{sec:acknowledge}
This research is supported in part by NSF grants CNS-1815959, CPS-2038923, III-2107200, and CPS-2038658.

\bibliography{mural}

% \begin{IEEEbiography}{Ahmet Soyyigit}
% Ahmet Soyyiğit received the B.S. and M.S. degrees in computer engineering from Gebze Technical University, Kocaeli, Türkiye, in 2016 and 2019, respectively, and the Ph.D. degree in computer science from the University of Kansas, Lawrence, KS, USA, in 2025, with a dissertation focused on anytime computing techniques for LiDAR-based perception in cyber-physical systems. He is currently an Assistant Professor with the Department of Computer Engineering, Turkish Naval Academy, National Defense University of Türkiye. His research interests include real-time embedded systems, deep neural networks, machine perception, anytime computing, robotics, and simulation.
% \end{IEEEbiography}

% \begin{IEEEbiography}{Shuochao Yao}
% \end{IEEEbiography}

%\newpage

% \begin{IEEEbiography}{Heechul Yun}
% \end{IEEEbiography}
\vspace{-3em}
\begin{IEEEbiographynophoto}{Ahmet Soyyiğit}
is an Assistant Professor of Computer Engineering at the National Defense University, Turkish Naval Academy. His research interests are in real-time embedded systems and artificial intelligence. He received a PhD in Computer Science from the University of Kansas.
\end{IEEEbiographynophoto}
% \vspace{-2cm}
\vspace{-3em}
\begin{IEEEbiographynophoto}{Shuochao Yao}
is an Assistant Professor of Computer Science at George Mason University. His research interests are in intelligent internet-of-things and cyber-physical systems. He received a PhD in Computer Science from the University of Illinois Urbana-Champaign.
\end{IEEEbiographynophoto}
\vspace{-3em}
\begin{IEEEbiographynophoto}{Heechul Yun}
is a Deane E. Ackers Scholar and Professor in the Department of Electrical Engineering and Computer Science at the University of Kansas. He leads research on safe and secure real-time computing infrastructure—spanning operating systems, computer architecture, and middleware—for intelligent cyber-physical systems. His work encompasses predictable real-time computing, embedded AI/ML, and hardware security. He has received multiple best and outstanding paper awards at top real-time systems venues. He earned his Ph.D. in Computer Science from the University of Illinois at Urbana–Champaign and previously worked at Samsung Electronics and ETRI.
\end{IEEEbiographynophoto}

% \clearpage

\end{document}